\documentclass[a4paper,twoside]{article}
\usepackage{epsfig}
\usepackage{subcaption}
\usepackage{calc}
\usepackage{amssymb}
\usepackage{amstext}
\usepackage{amsmath}
\usepackage{amsthm}
\usepackage{multicol}
\usepackage{pslatex}
\usepackage{apalike}
\usepackage{algorithm2e}
\usepackage[bottom]{footmisc}
\usepackage{SCITEPRESS}     
\usepackage{hyperref}
\begin{document}

\title{Predicting Depth Maps from Single RGB Images and Addressing Missing Information in Depth Estimation}

\author{\authorname{Mohamad Mofeed Chaar\orcidAuthor{0000-0001-9637-5832}, Jamal Raiyn\orcidAuthor{0000-0002-8609-3935} and Galia Weidl\orcidAuthor{0000-0002-6934-6347}}
\affiliation{Connected urban mobility, Faculty of Engineering,\\ University of Applied Sciences, Aschaffenburg, Germany}
\email{\{MohamadMofeed.Chaar, Jamal.Raiyn, Galia.Weidl\}@th-ab.de}
}


\abstract{Depth imaging is a crucial area in Autonomous Driving Systems (ADS), as it plays a key role in detecting and measuring objects in the vehicle’s surroundings. However, a significant challenge in this domain arises from missing information in Depth images, where certain points are not measurable due to gaps or inconsistencies in pixel data. Our research addresses two key tasks to overcome this challenge. First, we developed an algorithm using a multi-layered training approach to generate Depth images from a single RGB image. Second, we addressed the issue of missing information in Depth images by applying our algorithm to rectify these gaps, resulting in Depth images with complete and accurate data. We further tested our algorithm on the Cityscapes dataset and successfully resolved the missing information in its Depth images, demonstrating the effectiveness of our approach in real-world urban environments.
}

\onecolumn \maketitle \normalsize \setcounter{footnote}{0} \vfill
\section{INTRODUCTION}\label{sec:introduction}
\begin{figure}
    \centering
    \includegraphics[width=1\linewidth]{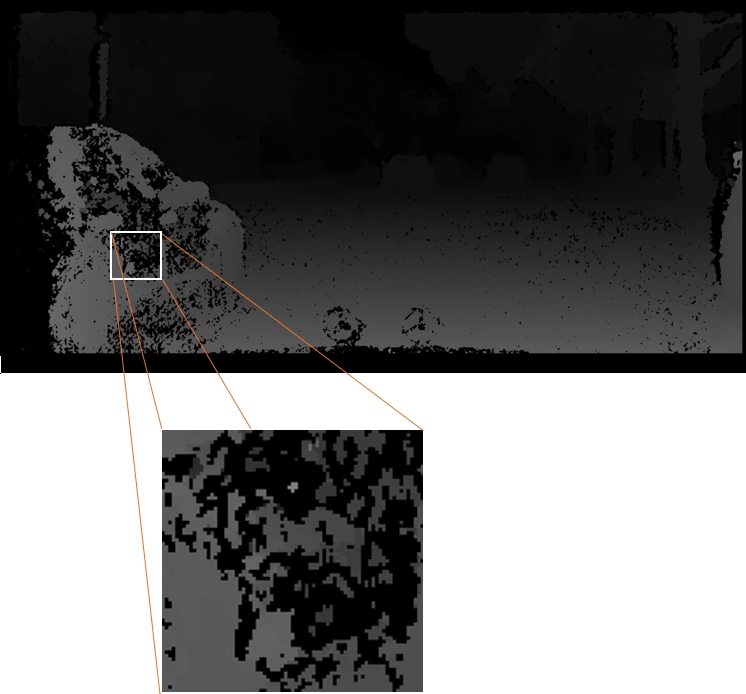}
     \put(-90,50.5){The black pixels represent} 
     \put(-90,40.5){invalid measurements }
    \caption{The Depth image from the Cityscape dataset contains black pixels, indicating missing information that affects the accuracy of the Depth image.}
    \label{fig:fig01}
\end{figure}

\begin{figure}[h]
    \centering
    \begin{subfigure}{.25\textwidth}
    \centering
    \includegraphics[width=.95\linewidth]{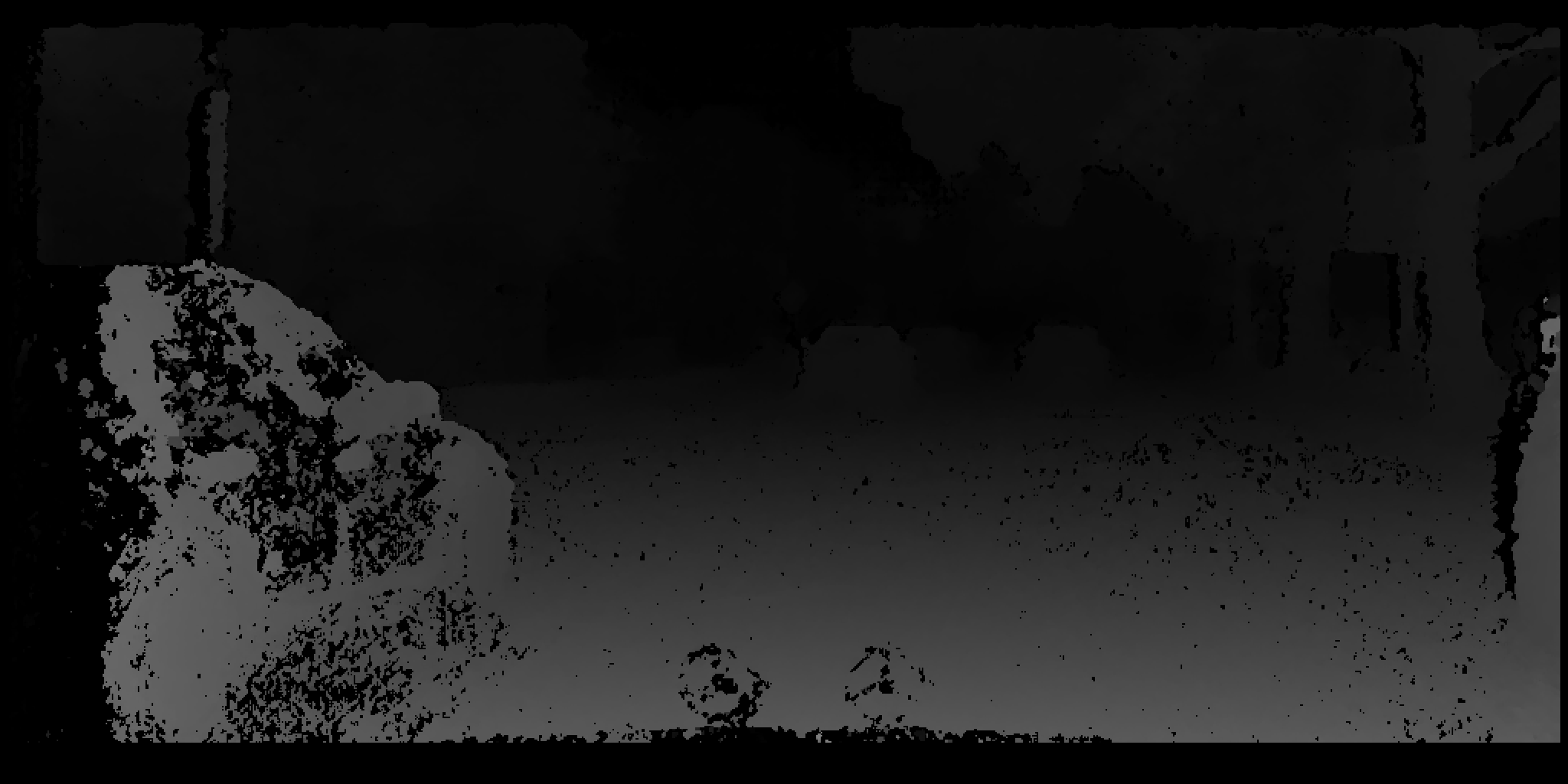}
    \caption{}
    \label{fig:sub1}
    \end{subfigure}%
    \begin{subfigure}{.25\textwidth}
    \centering
    \includegraphics[width=.95\linewidth]{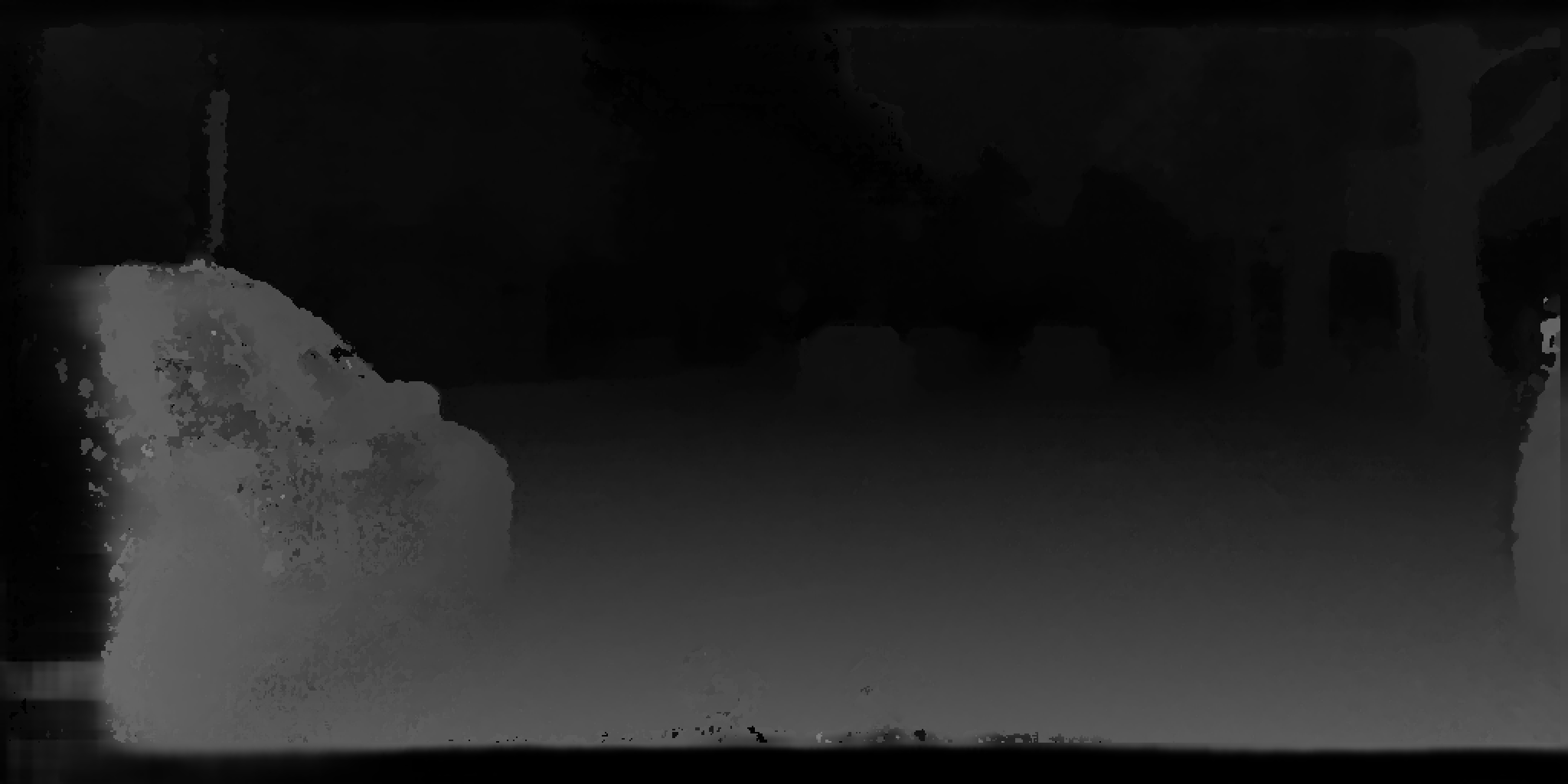}
    \caption{}
    \label{fig:sub2}
    \end{subfigure}
    
    \caption{Image (a) shows an example from the Cityscapes dataset before applying our filtering technique to address missing pixels. Image (b) shows the result after applying our method, where the missing information has been successfully restored.}
    \label{fig:fig02}
\end{figure}

Depth images are used to estimate the distance of objects at each pixel by employing a stereo camera. The Cityscapes dataset \cite{Ref001,Ref002} Depth images are used to estimate the distance of objects at each pixel by employing a stereo camera. The Cityscapes dataset (Figure\ref{fig:fig01}). However, a significant challenge in Depth imaging is the presence of missing information, which is a known issue in the Cityscapes dataset. 
Our work focuses on two key tasks:
\begin{itemize}
    \item Generating Depth images from a single RGB image using a multi-level training approach based on an image segmentation algorithm. 
    \item Enhancing the accuracy of Depth images by correcting missing information through our proposed methodology, achieving an accuracy rate of 90.19\%. 
\end{itemize}
In addition, our work enhanced the accuracy of Depth image generation from a single RGB image through image segmentation\cite{Ref003,Ref004}. \\
In a separate phase, we employed deep learning techniques to generate Depth images from the Cityscapes dataset, followed by applying our model to refine the data further. This refined dataset has been then used as a target to train a new model, continuing an iterative process (loop training) that has progressively improved the Cityscapes dataset and has resulted in higher accuracy (Figure \ref{fig:fig02}). In this study, we have applied the loop training process five times. After the first training iteration, the accuracy of generating Depth images reached 82.78\%. In the last training loop, the accuracy increased to 90.19\%, representing an improvement of approximately 7.41\%.

\bigskip

\section{RELATED WORK}\label{sec:RELATED WORK}
 Monocular Depth estimation is the task of predicting a Depth map (distance from the camera) from a single RGB image. This is a challenging problem due to the lack of direct Depth cues (like stereo vision or motion) in a single image. 
 Moreover, Many Depth estimation models trained on specific datasets tend to perform well on those datasets but fail when tested on new or unseen datasets due to differences in lighting conditions, camera characteristics, or environmental contexts. This is a significant issue because creating large-scale annotated Depth datasets is resource-intensive. The paper \cite{Ref005} mixed the data for better generalization, and the severe weather \cite{Ref007,Ref008} has also been included in their data. The authors \cite{Ref006} introduced a method for Depth completion, converting sparse LiDAR data into dense Depth maps using RGB images without needing ground-truth Depth annotations. The approach leveraged self-supervised learning through photometric and geometric consistency losses, enabling the model to train effectively by using sparse LiDAR points and monocular images. It also incorporated stereo image pairs to enhance Depth accuracy. This framework achieved robust, high-quality Depth estimation, providing a practical solution for applications like autonomous driving, this methodology works well in specific scenes that are used in training, but it is limited when we generalize it in other data.
The paper \cite{Ref009} approached predicting relative Depth from a single image using stereo data from the web for weak supervision. Instead of relying on costly ground-truth Depth annotations, the method leveraged geometric cues from stereo images (3600 images) to train a model that can generalize across diverse scenes. This scalable approach pushes the boundaries of monocular Depth estimation, making it effective for real-world applications without requiring extensive labeled data.
The paper \cite{Ref010} introduced a monocular Depth prediction using a novel skip attention mechanism. The model leveraged attention layers focusing on important features across different network scales. By incorporating skip connections with attention, the model efficiently combines high-level semantic information with low-level spatial details, improving Depth prediction accuracy. The authors \cite{Ref011} present an approach for predicting Depth maps from a single RGB image by utilizing a multi-scale deep network. The model processes images at different scales, capturing both global scene context and fine-grained details to improve Depth estimation accuracy. By combining features from multiple resolutions, the network is able to predict Depth for complex scenes better, handling variations in scale and structure more effectively. 
The paper \cite{Ref012} introduced a novel approach for monocular Depth estimation using a multi-scale Laplacian pyramid to fuse Depth predictions at different scales. By capturing both global and local Depth information, the model refines Depth estimates through residual learning at each pyramid level.

Previous studies primarily focused on generating Depth images by utilizing Depth images as the target output. In contrast, our methodology has introduced improvements to the datasets by addressing and correcting missing pixels, allowing us to use the refined Depth images as target outputs. This approach resulted in a significant increase in accuracy, achieving a performance of 90.19\% as we will see later.

\bigskip

\section{Methodology}\label{sec:Methodology}

\subsection{Image Segmentation}
\begin{figure*}
    \centering
    
    \includegraphics[width=1\linewidth]{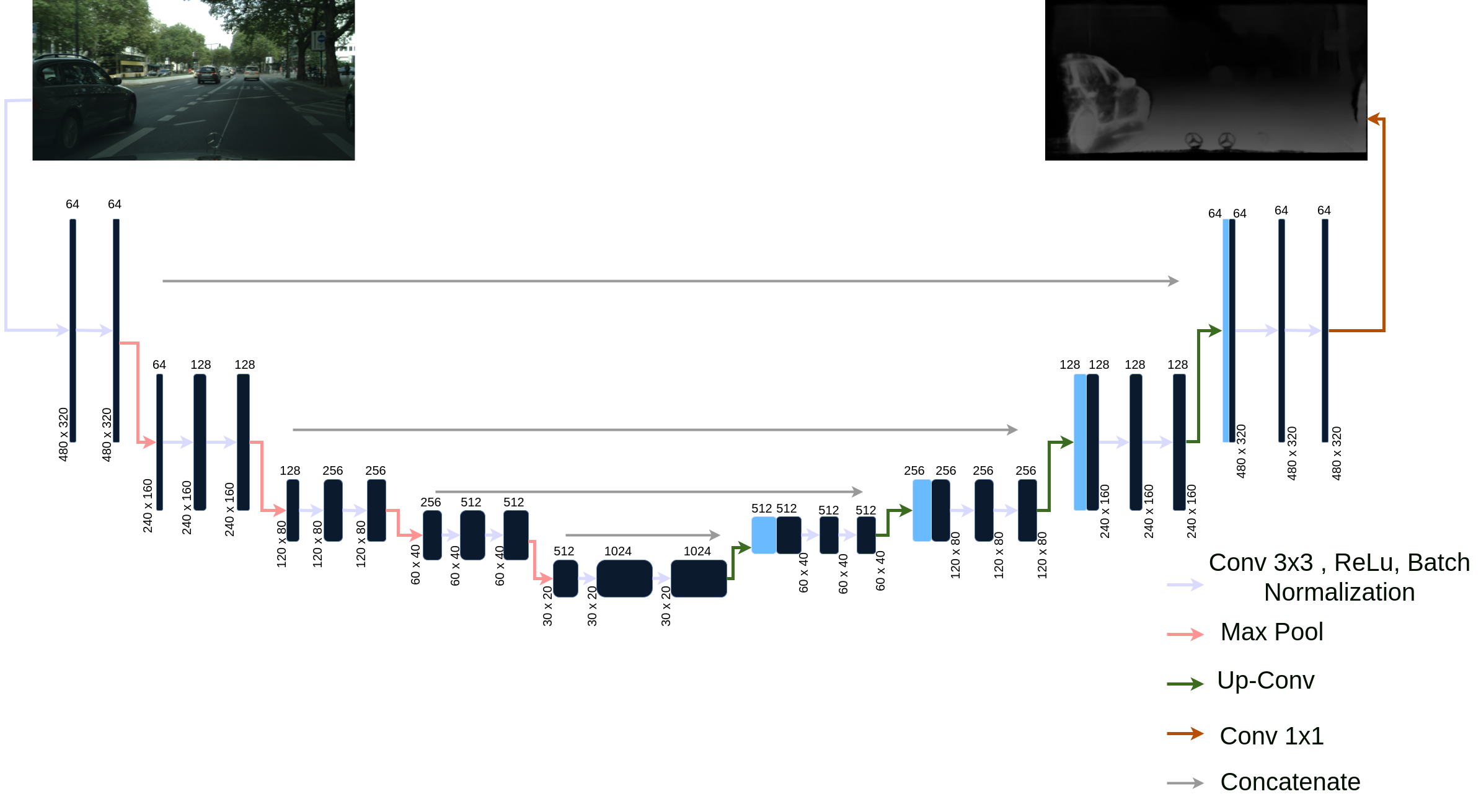}

    \caption{Creator of the U-Net algorithm, which was implemented in our work.}
    \label{fig:fig03}
\end{figure*}
Image segmentation is the process of partitioning an image into multiple segments or regions for each pixel, with the goal of simplifying the representation of an image to make it more meaningful and easier to analyze. One of the most effective architectures for image segmentation tasks, especially in medical imaging, is the U-Net architecture \cite{Ref013}, which operates the encoder and decoder layers to deduce the segmentation for each pixel.
In practice, U-Net can be trained using annotated images where each pixel is labeled as belonging to a specific class or object. The loss function that we used in this work is the Mean Square Error Loss function (MSELoss).
In our work, we used U-Net to train on the Cityscapes dataset, with the Depth image as the target mask. Both input and output images have a resolution of 480x320 pixels (Figure \ref{fig:fig03}). 

The stereo camera provides two images (left and right) for each depth image, but since our goal is to generate depth from a single image, we selected only the left image for training and omitted the right image.

\bigskip
\subsection{Metrics}\label{subsec:Metrics}
\subsubsection{Accuracy}\label{aubsubsec:Accuracy}
The accuracy in this work is evaluated based on the Absolute Error between the predicted pixel values and the target values of the depth image, calculated using the following formula:
\begin{equation}\label{equ:02}
Aboolute\ Error\ pixels = \sum_{P \in Pixels}|P-\hat{P}| 
\end{equation}
Where $P$  represents the actual pixel value, and $\hat{P}$ denotes the predicted pixel value, the percentage of accuracy is calculated by dividing the absolute error of the pixels by the sum of all pixel values in the image, as expressed in the following formula:
\begin{equation}\label{equ:03}
Accuracy = \frac{Aboolute\ Error\ pixels}{\sum_{P \in Pixels}{P}} *100
\end{equation}
\subsubsection{Corrected Pixels}\label{aubsubsec:Corrected Pixels}
The primary objective of this work is to correct missing information in depth images, represented as black pixels. To establish a criterion for addressing this issue, we calculated the average number of black pixels per image (\nameref{Appendix A}), which was found to be 1,206,898. To validate our approach, we calculated the average number of corrected black pixels per iteration, using this as a performance criterion, as expressed in the following formula which calculate the percentage of corrected pixels:

\begin{equation}\label{equ:04}
Corrected\ Pixels = \frac{Average\ of\ corrected\ pixels}{Average\ black\ pixels}*100
\end{equation}

\bigskip
\subsection{Depth image}\label{subsec:Depth image}
Typically, generating a Depth image dataset involves using a stereo camera\cite{Ref016} setup, where two cameras capture images from slightly different angles. The Depth information is obtained by calculating the disparity between corresponding points in the two images.

\begin{figure}[hb]
    \centering
    
    \includegraphics[width=1\linewidth]{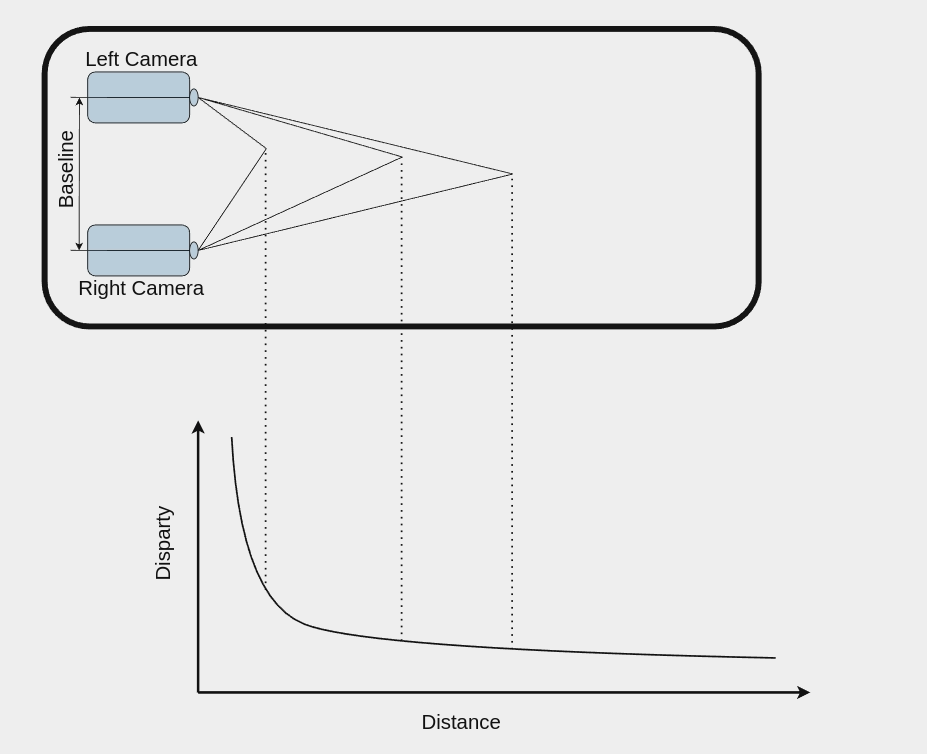}
     
    \caption{The disparity value can be used to calculate distance using Equation \ref{equ:01}}
    \label{fig:fig04}
\end{figure}

Estimate Depth of pixels is calculated by the following equation\cite{Ref014}:
\begin{equation}\label{equ:01}
    Z = Bf/disparity
\end{equation}
Where $B$ (baseline) is the distance between two cameras where it is 22 cm in the Cityscapes dataset\cite{Ref001}, $Z$ is the Depth of the pixel, and $f$ is the focal length.\\
In cityscape datasets,  $disparity$ calculated by the following\cite{Ref015}: 
\begin{equation}
disparity = ( float(p) - 1. ) / 256
\end{equation}
Where $p$ is the value of a pixel in the Depth image, and it is between [0,126]; furthermore, $p = 0$ when the measurement is invalid (Figure \ref{fig:fig04}) \cite{Ref017}.

\bigskip

\subsection{Cityscape Datasets}\label{subsec:Cityscape Datasets}
\begin{figure}
    \centering
    
    \includegraphics[width=1\linewidth]{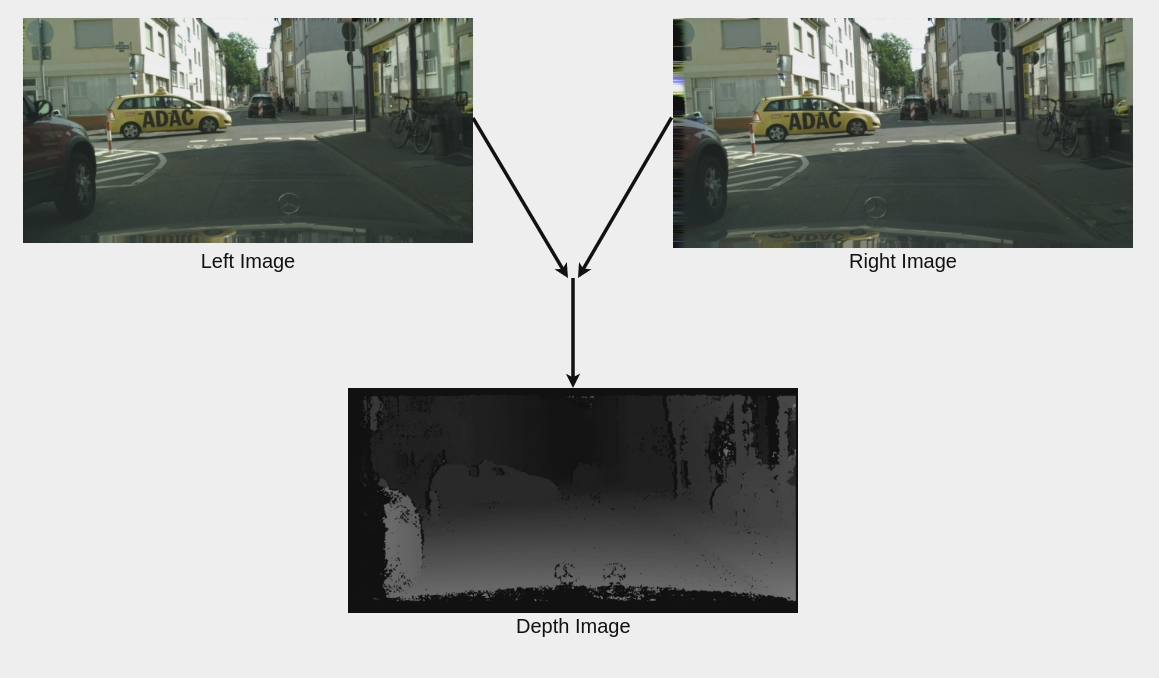}
     
    \caption{The image was captured from the Cityscapes dataset in Frankfurt. Using the left and right stereo images, a depth image can be generated.}
    \label{fig:fig05}
\end{figure}

The Cityscapes dataset \cite{Ref001,Ref002,Ref018} is a large-scale dataset created for urban scene understanding, particularly tailored for research in semantic segmentation, instance segmentation, and depth prediction in complex city environments. This dataset is widely used in computer vision research due to its high-quality annotations and realistic street scenes captured from various German cities.
The depth images in the Cityscapes dataset were generated using stereo camera sensors mounted on a vehicle as it drove through various European cities. The dataset is organized into three folders: train, test, and validation, containing a total of approximately 150,000 images with size (2048x1024) pixels. In addition to the depth images, corresponding left and right camera images are provided, enabling comprehensive analysis and training for depth estimation and other vision-based tasks.\\
As shown in Figure \ref{fig:fig05}, the depth image contains black points, representing invalid measurement areas. In the Cityscapes dataset, each image has an average of 1,206,898 black pixels (\nameref{Appendix A}), accounting for 57.5\% of the image, indicating missing information, which our work is concerned with solving this problem.

\bigskip

\subsection{Iterative training}\label{subsec:Iterative training}
\begin{figure}[ht]
    \centering
    
    \includegraphics[width=1\linewidth]{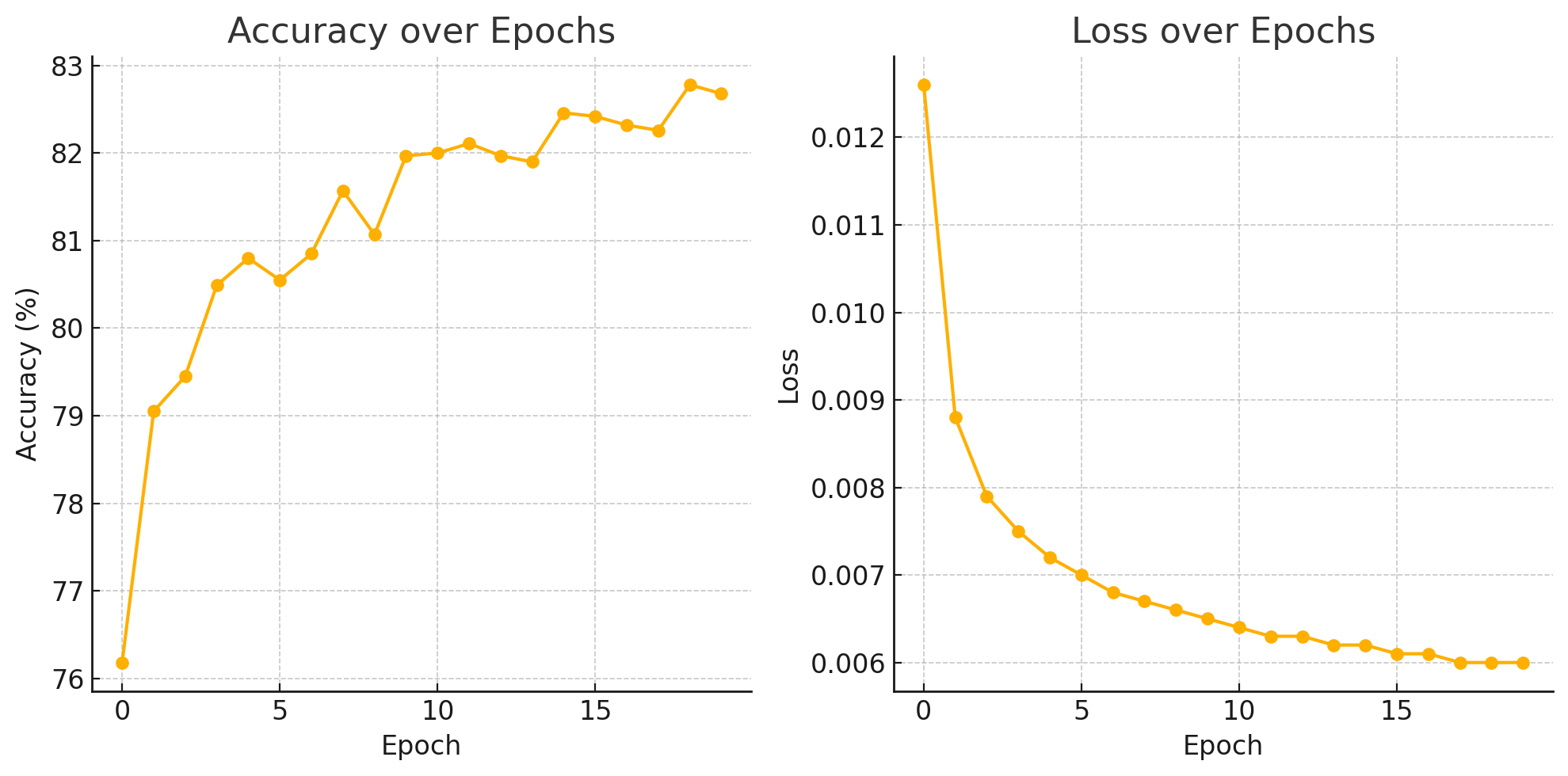}

    \caption{the graphs showing how accuracy and loss change over the steps. The left graph displays the accuracy trend, while the right graph shows the loss trend across each step; the highest accuracy is 82.78\%.}
    \label{fig:fig06}
\end{figure}

The missing information impacts the accuracy of the training data, reducing accuracy when this data is used as a target mask in image segmentation algorithms, resulting in an accuracy of 82.78\% (Figure \ref{fig:fig06})

The image segmentation results after training the data give us a result without missed information which means a smoother image but with an accuracy of 82.78\% for the pixels that have been filled in Depth image (Figure \ref{fig:fig07}).\\ 
We trained our model for image segmentation to enhance the Depth of images and address the issue of black pixels. After predicting each image, we examined every pixel in the Depth image from the Cityscape dataset. The zero-value pixels are replaced with the corresponding pixel values predicted by our model for the same location, effectively filling in the missing Depth information (Algorithm \ref{algorithm:1})(Figure \ref{fig:fig08}).

\begin{algorithm}
 \caption{How to Fix black pixels in Depth image}
 \KwData{Depth image}
 \KwResult{Fix missing information in Depth image}
 initialization\;
 predict depth image using our model\;
 \While{There is Pixel with value zero}{
  Replace it with the value of the predicted image in the same position\;
  }\label{algorithm:1}
\end{algorithm}

\begin{figure}
    \centering
    \begin{subfigure}{.25\textwidth}
    \centering
    \includegraphics[width=.95\linewidth]{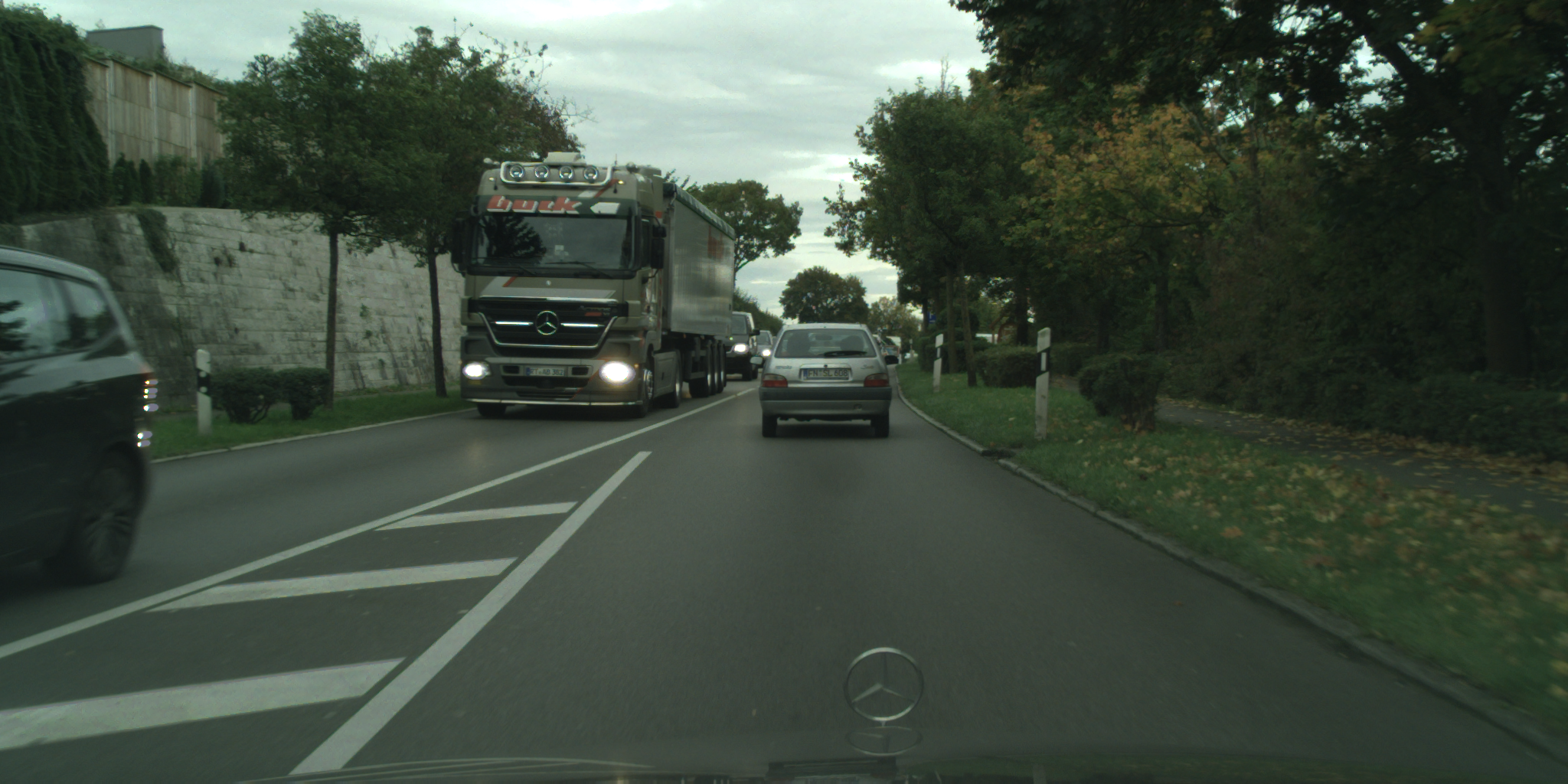}
    \caption{}
    \label{fig:sub7_1}
    \end{subfigure}%
    \begin{subfigure}{.25\textwidth}
    \centering
    \includegraphics[width=.95\linewidth]{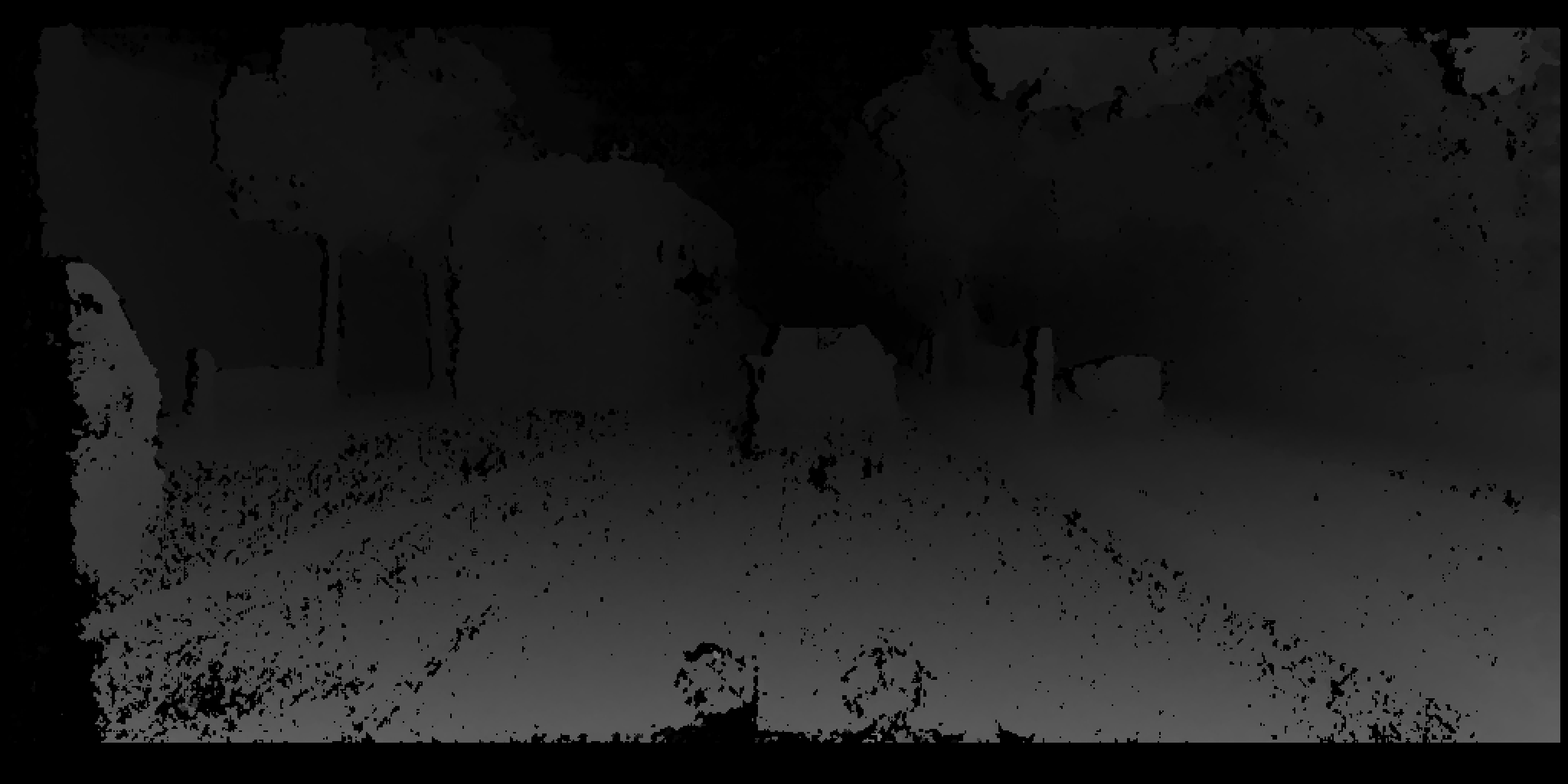}
    \caption{}
    \label{fig:sub7_2}
    \end{subfigure}
    
    \begin{subfigure}{.25\textwidth}
    \centering
    \includegraphics[width=.95\linewidth]{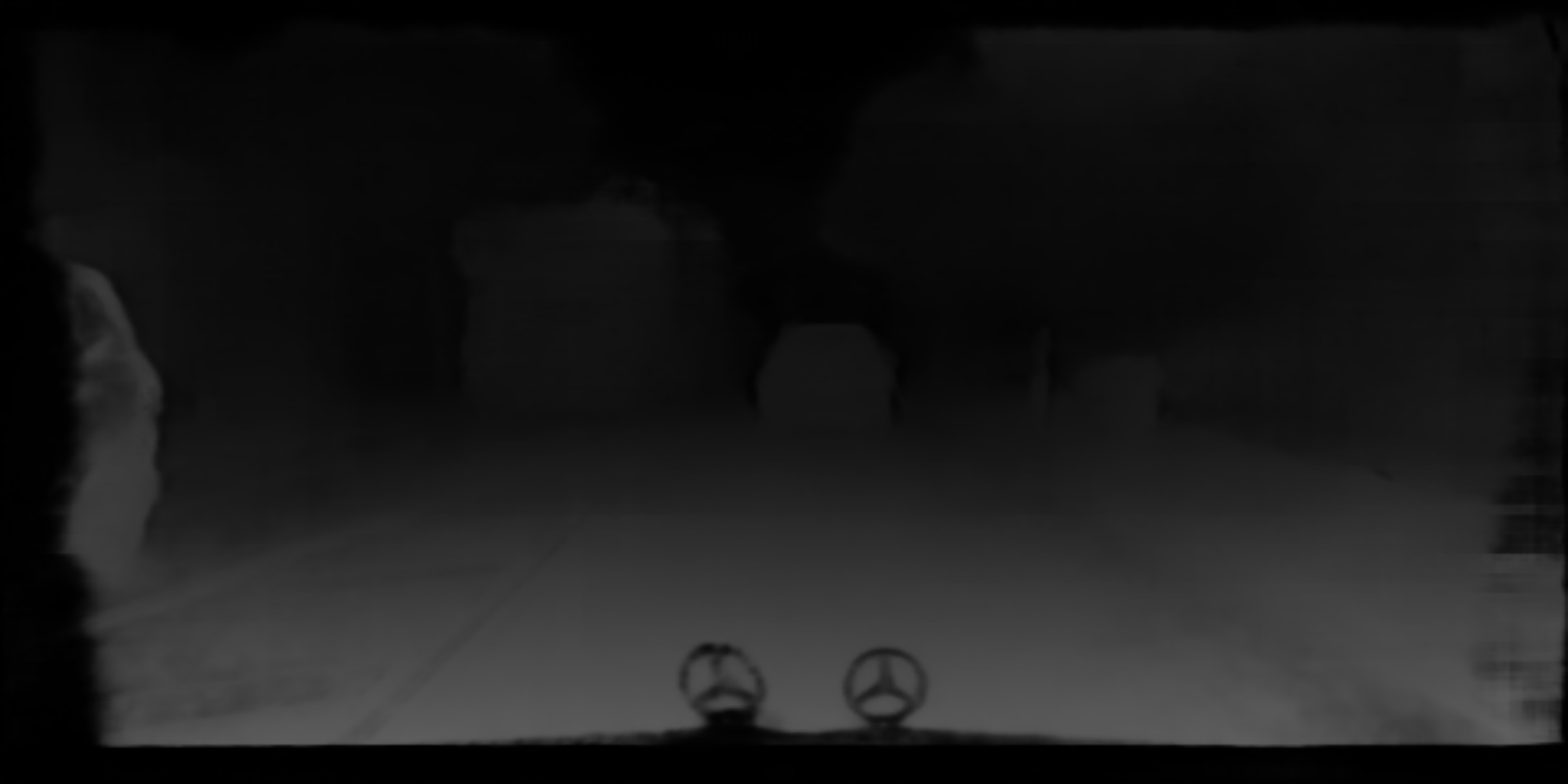}
    \caption{}
    \label{fig:sub7_3}
    \end{subfigure}%
    \begin{subfigure}{.25\textwidth}
    \centering
    \includegraphics[width=.95\linewidth]{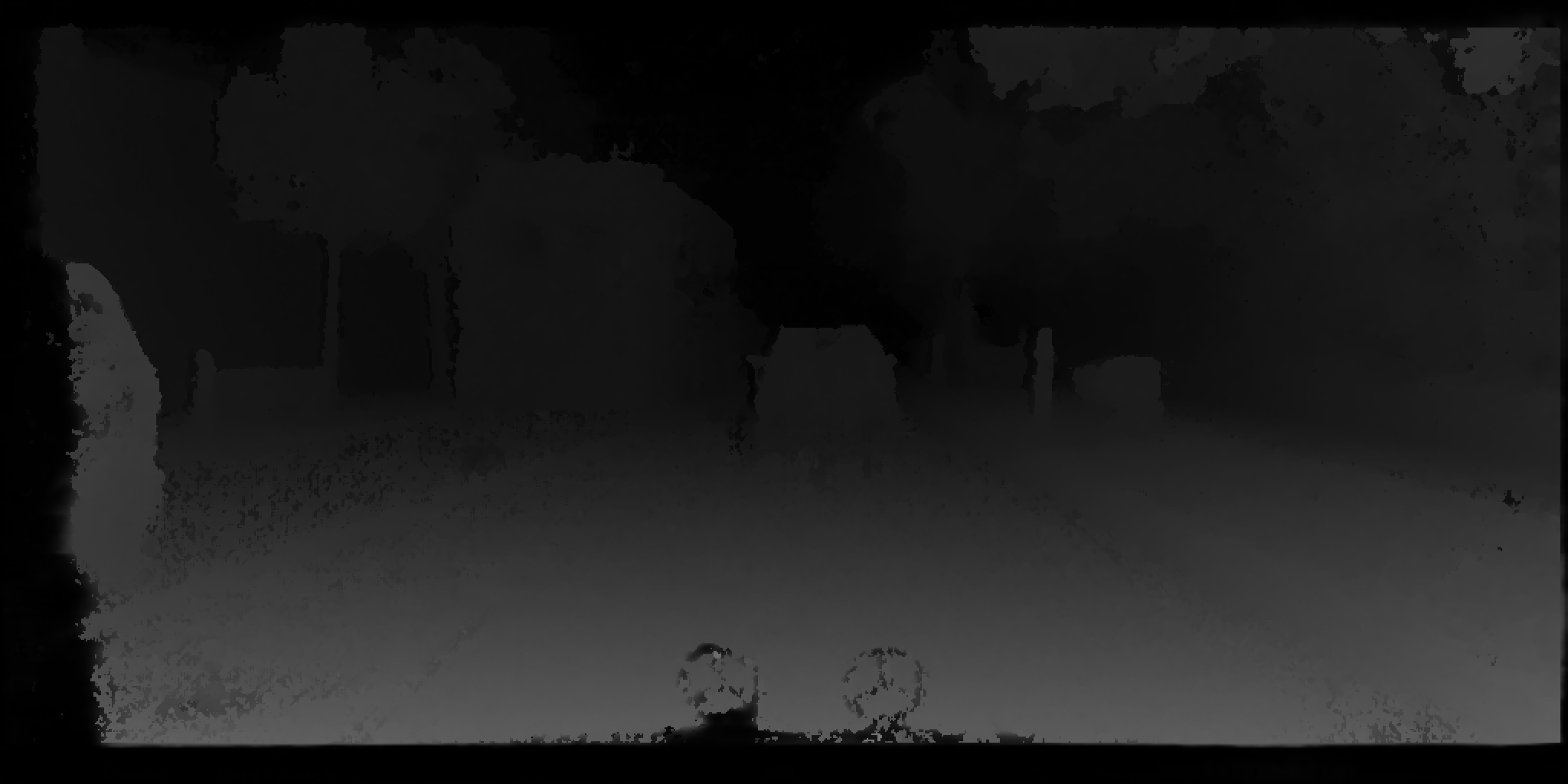}
    \caption{}
    \label{fig:sub7_4}
    \end{subfigure}
    
    \caption{Figure (a) presents an example from the Cityscapes dataset captured in Lindau city. Figure (b) depicts a Depth map generated using the stereo camera. Figure (c) illustrates the predicted depth map produced by our initial model training. Finally, Figure (d) demonstrates an enhanced version of the depth map generated by the stereo camera (Figure b), achieved through our proposed method, as reflected in the improvements shown in Figure (c).}
    \label{fig:fig07}
\end{figure}
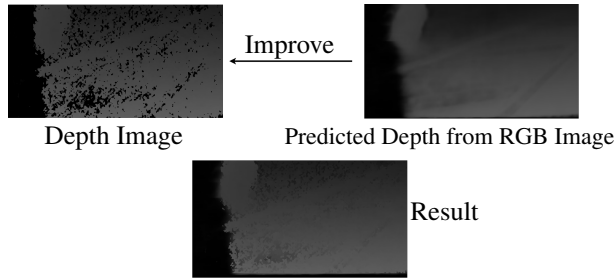
\begin{figure}
    \centering
    \includegraphics[width=1\linewidth]{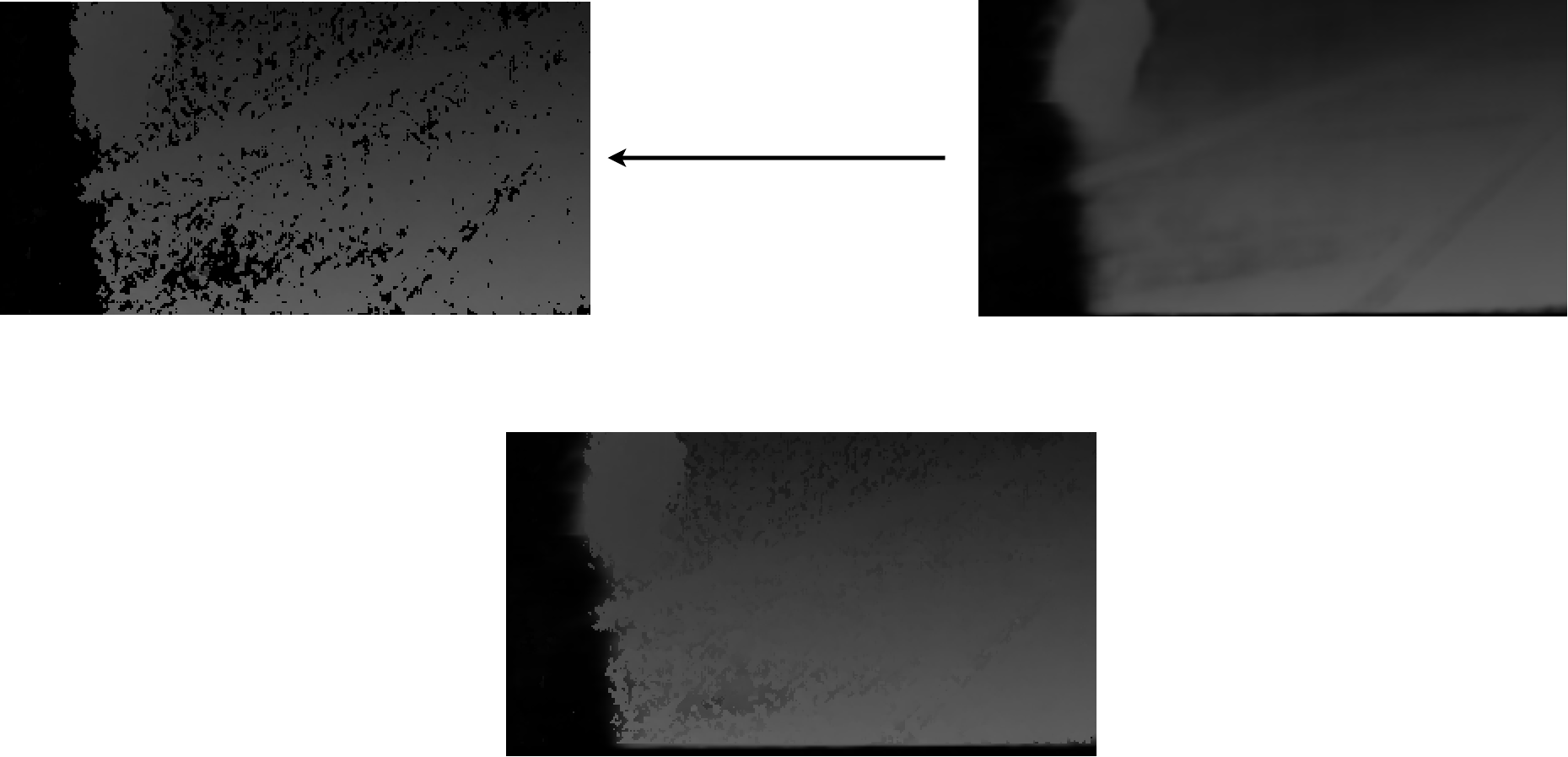}
     \put(-200,50.5){Depth Image} 
     \put(-110,50.5){\small Predicted Depth from RGB Image}
     \put(-125,85){Improve}
     \put(-63,22){Result}
    \caption{The Depth image from the Cityscape dataset contains black pixels, indicating missing information that affects the accuracy of the Depth image.}
    \label{fig:fig08}
\end{figure}

\begin{figure}
    \centering
    \includegraphics[width=1\linewidth]{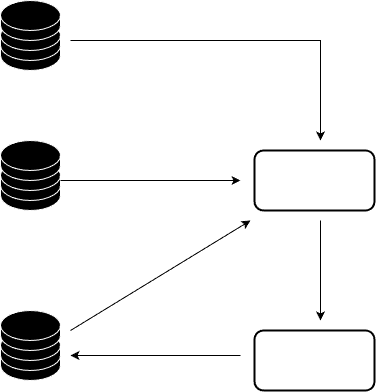}
     \put(-215,225){RGB Images} 
     \put(-215,148){Depth Images} 
     \put(-125,125){Train}
     \put(-125,80){Train}
     \put(-60,120){New Model}
     \put(-90,60.5){Predict Depth}
     \put(-80,50.5){images}
     \put(-61,25){\small Data Predicted}
     \put(-56,15){\small from the last}
     \put(-46,5){\small model}
     \put(-165,25){\small Improve Depth images}
     \put(-215,60){\small Enhanced}
     \put(-215,50.5){\small Depth Images}
    \caption{This diagram illustrates the iterative training process, where we train our datasets to predict depth images. We then enhance the depth images in the Cityscape dataset and retrain using the enhanced images. With each iteration, we observe an increase in accuracy. In this work, we repeated the training process five times.}
    \label{fig:fig09}
\end{figure}
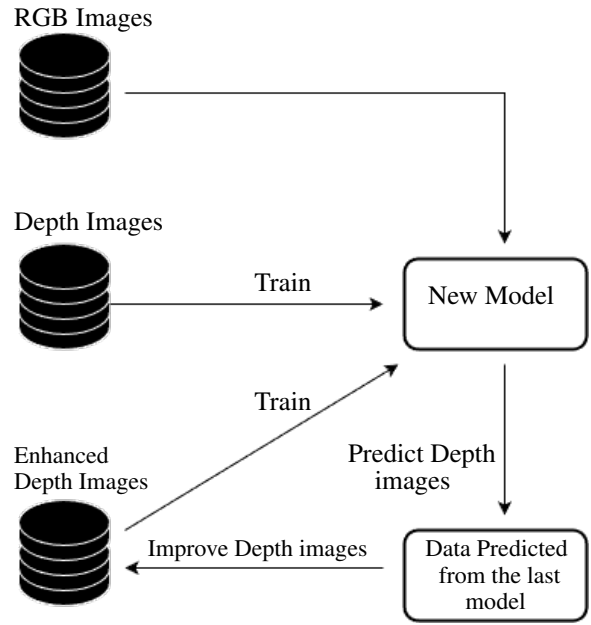

The results shown in Figures \ref{fig:fig07} and \ref{fig:fig08} illustrate our initial training (first iteration). At this stage, we enhanced the Depth images, achieving complete images without missing information, which we filled using our predictions with an accuracy of 82.78\%. These improved images can now be utilized to train our data further and serve as masks for image segmentation in the next training iteration. This approach increases accuracy, as the masks used are free from missing information (Algorithm \ref{algorithm:2}) (Figure \ref{fig:fig09}).

\begin{algorithm}
 \caption{In iterative training, we process the datasets using the latest depth images, which we have enhanced.}
 \KwData{Depth image}
 \KwResult{Depth image without missed information}
 
 \While{Iteration}{
  Train a new model using Depth images as a mask\;
  Improve Depth image using our model (Algorithm \ref{algorithm:1})\;
  Update Depth images by improved Depth images\;
  }\label{algorithm:2}
\end{algorithm}

When using enhanced depth images, the accuracy of training the dataset improved with each iteration. During the first training iteration, an accuracy of 82.78\% was achieved. After enhancing the datasets, the accuracy increased to 88.16\%. Based on this improvement, we decided to iterate the training process five times, ultimately achieving an accuracy of 90.19\%.

\begin{table}
\caption{The accuracy and the percentage of corrected pixels improved significantly through our iterative learning process. As observed, the training accuracy of our model increased steadily, and the percentage of corrected black pixels also showed a consistent rise. }
\label{tab:tab2} 
\centering
\begin{tabular}{|c|c|c|}
  \hline
  Iterative &  $Accuracy^{1}$ & $Corrected \ Pixels^{2}$ \\
  \hline
  1 & 82.78\% & 58.3\%\\
  \hline
  2 & 88.16\% & 65.8\%\\
  \hline
  3 & 89.11\% & 70.5\%\\
  \hline
  4 & 89.83\% & 72,5\%\\
  \hline
  5 & 90.19\% & 73.3\%\\
  \hline
\end{tabular}
\footnotesize$^{1}${According to our formula at Equation \ref{equ:03}}\\
\footnotesize$^{2}${According to our formula at Equation  \ref{equ:04}}\\
\end{table}
\bigskip
\subsection{Correct Missed Information Using U\_Net}
\begin{figure}[hb]
    \centering
    \begin{subfigure}{.25\textwidth}
    \centering
    \includegraphics[width=.95\linewidth]{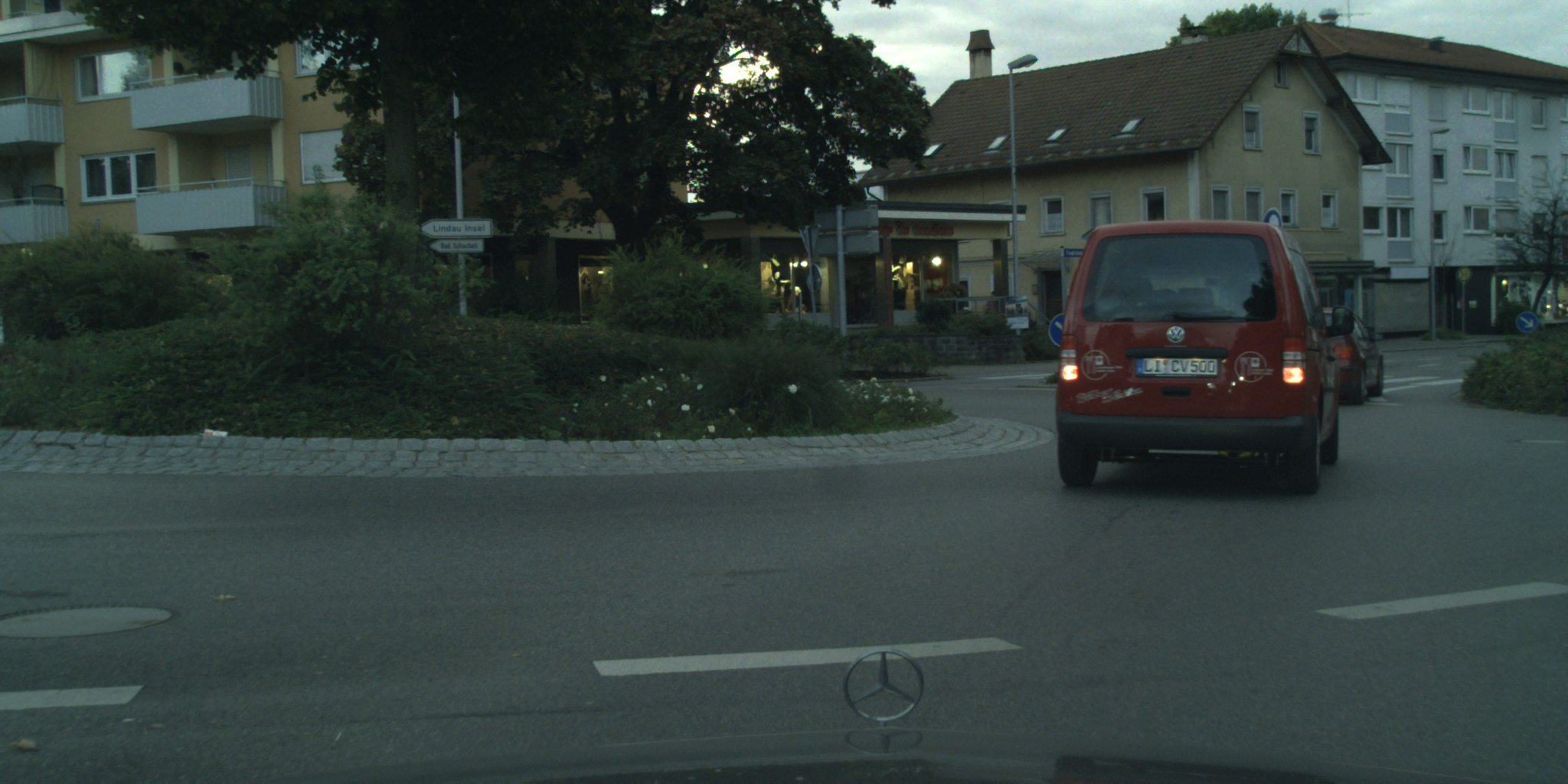} 
    \caption{}
    \label{fig:sub10_1}
    \end{subfigure}%
    \begin{subfigure}{.25\textwidth}
    \centering
    \includegraphics[width=.95\linewidth]{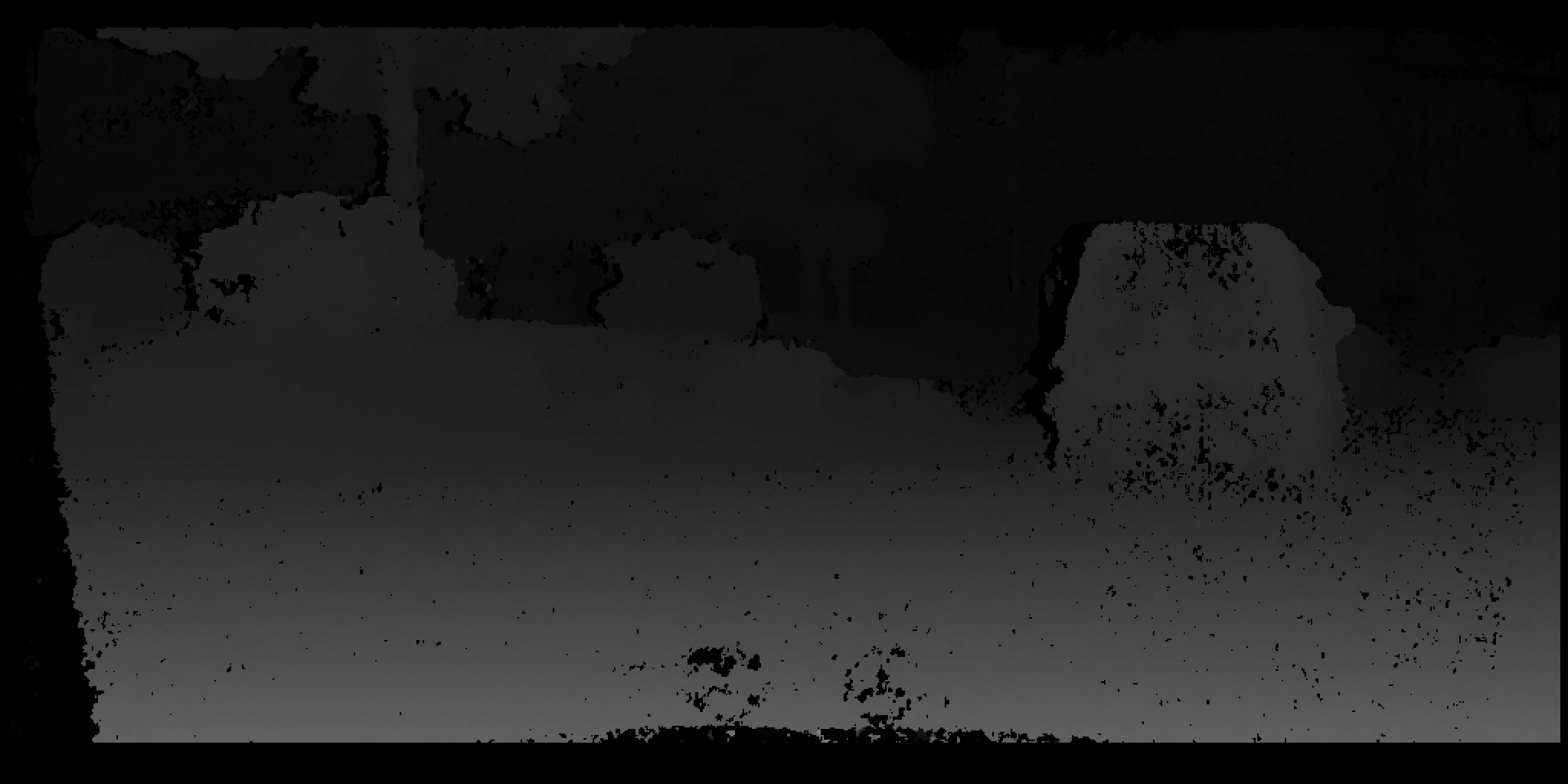} 
    \caption{}
    \label{fig:sub10_2}
    \end{subfigure}
    
    \begin{subfigure}{.25\textwidth}
    \centering
    \includegraphics[width=.95\linewidth]{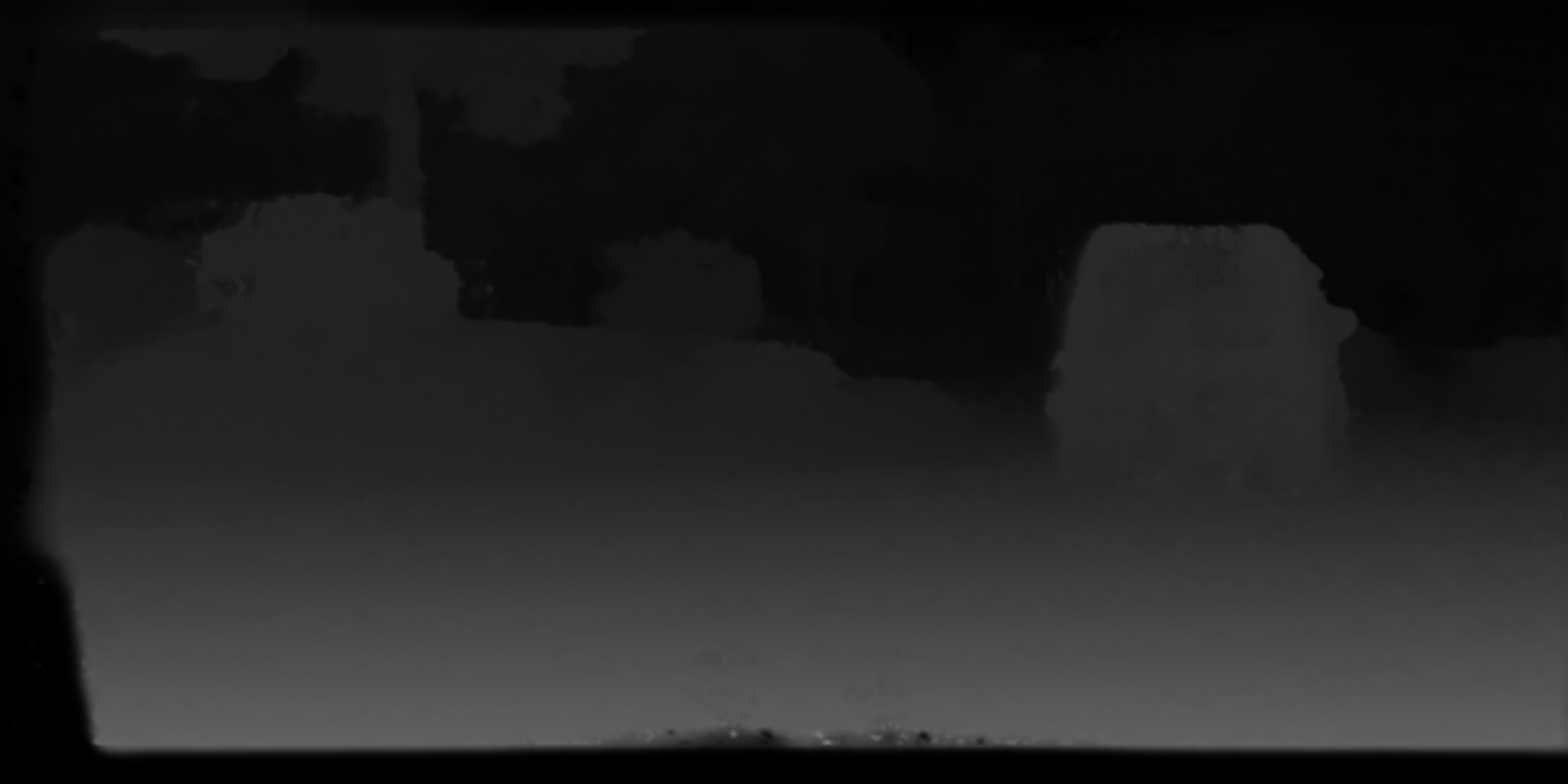} 
    \caption{}
    \label{fig:sub10_3}
    \end{subfigure}%
    \begin{subfigure}{.25\textwidth}
    \centering
    \includegraphics[width=.95\linewidth]{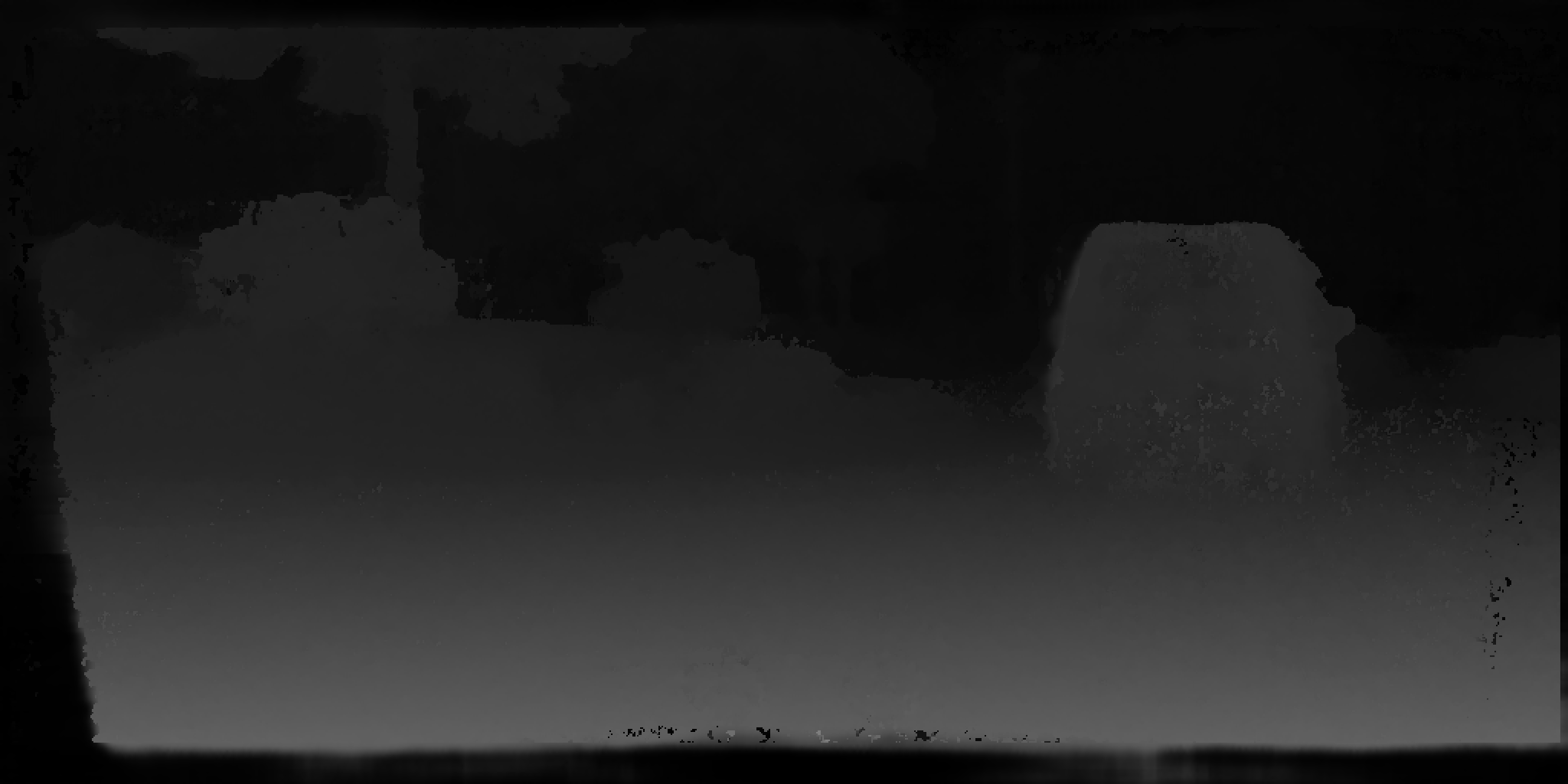} 
    \caption{}
    \label{fig:sub10_4}
    \end{subfigure}
    
    \caption{Figure (a) presents an example from the Cityscapes dataset captured in Lindau city. Figure (b) depicts a Depth map generated using the stereo camera. Figure (c) illustrates our model to enhance the missed information from the depth image using the U-Net algorithm (the input is an image (b)). Finally, Figure (d) demonstrates an enhanced version of the depth map generated by the stereo camera (Figure b), achieved through our proposed method, as reflected in the improvements shown in Figure (d).}
    \label{fig:fig10}
\end{figure}

According to our previous work, we improved depth images by filling in the missing information with highly accurate predictions during the final iteration of training (90.19\%). However, the correction process is time-consuming and impractical, as it requires predicting each image and checking every pixel to fill in the missing information (806 ms for each frame) for our computer properties, see \nameref{Appendix B}. Fortunately, we already have depth images and their corrected versions from our earlier work. This allows us to train a new image segmentation model, using the depth images with missing information as input and the corrected images as masks. The model can then predict new images with corrected missing information using AI, where we have achieved an accuracy of 96.44\% (Figure \ref{fig:sub10_4}) which it costs (19 ms for each frame).

\bigskip

\section{RESULTS AND DISCUSSION}\label{sec:Results and Discussions}
Building upon our previous approach, we successfully accomplished multiple tasks in this work, as outlined below:
\begin{itemize}
\item Generating depth images from a single RGB image using an iterative training approach, we steadily improved the accuracy of depth predictions with each iteration. The iterative process demonstrated consistent progress, with accuracy increasing at each step, showcasing the effectiveness of the refinement and optimization strategies applied during training.
\item Improving Depth Image Accuracy with U-Net and Self-Supervised Learning: Leveraging enhanced depth maps as masks to correct missing information, achieving 96.44\% accuracy.
\item 
\end{itemize}

\begin{figure}[hb]
    \centering
    \begin{subfigure}{.5\textwidth}
    \centering
    \includegraphics[width=.95\linewidth]{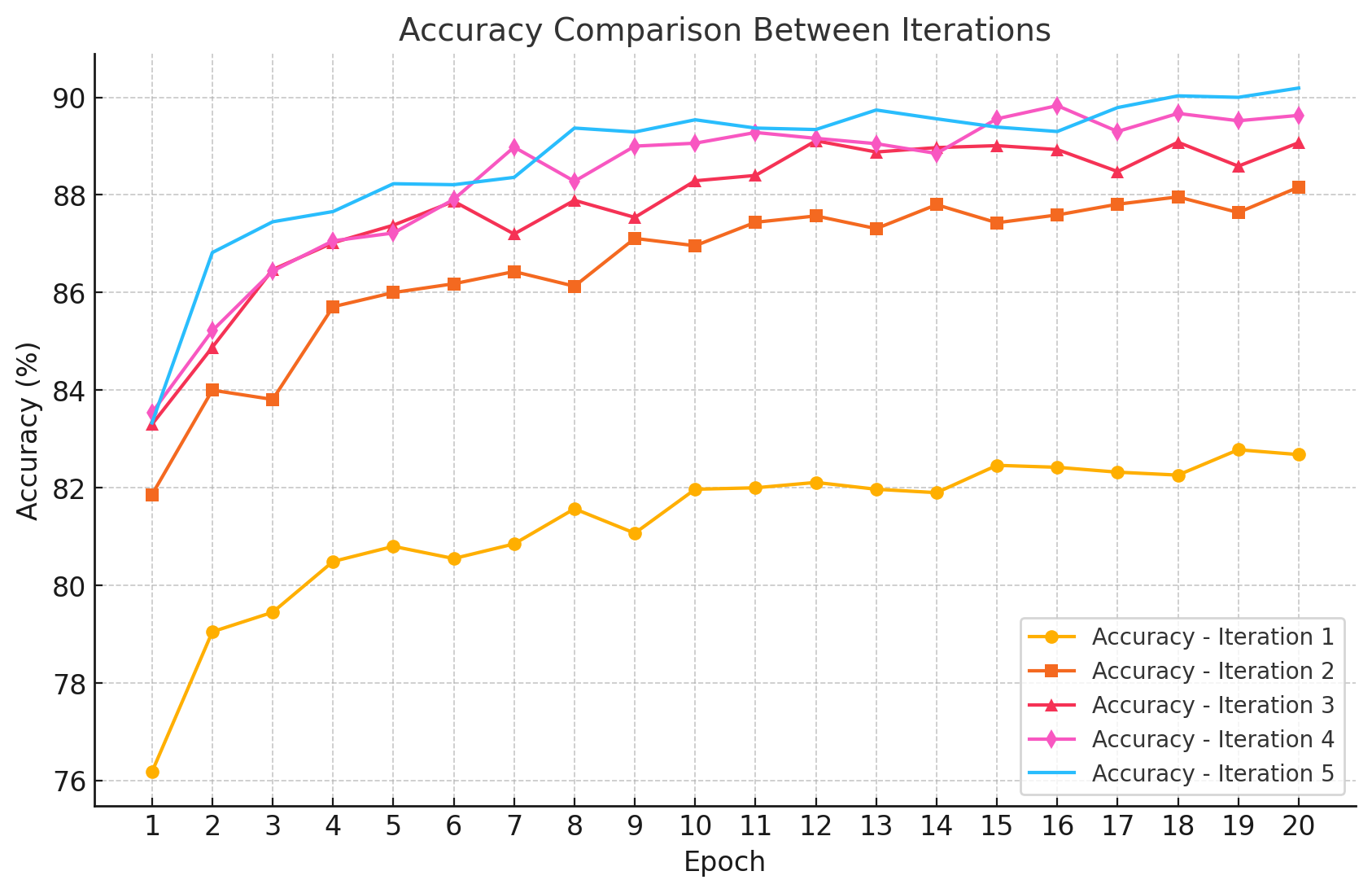}
    \caption{}
    \label{fig:sub11_1}%
    \end{subfigure}
    \begin{subfigure}{.5\textwidth}
    \centering
    \includegraphics[width=.95\linewidth]{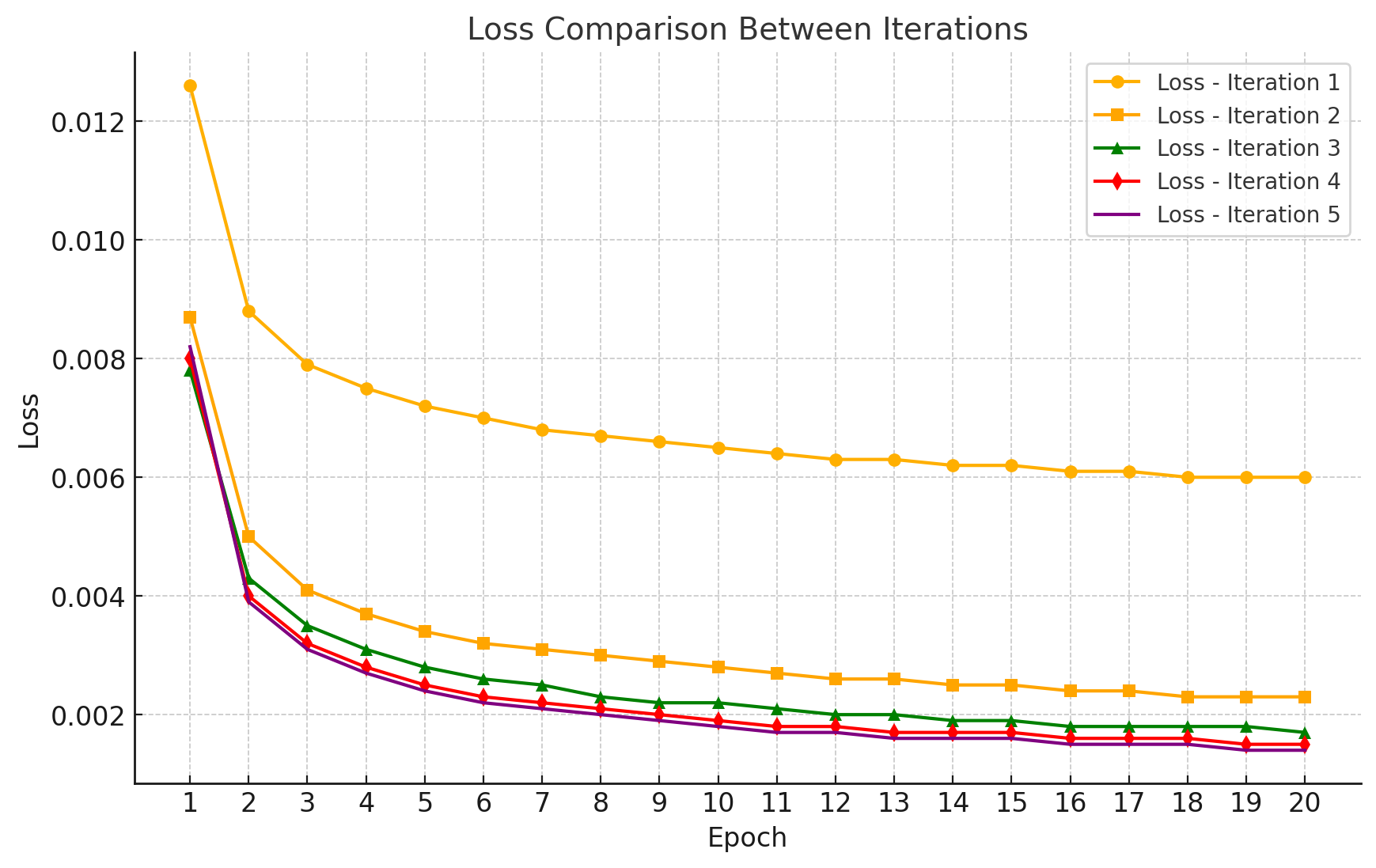}
    \caption{}
    \label{fig:sub11_2}
    \end{subfigure}

    \caption{Figure a: Accuracy Trends Across Epochs for Multiple Iterations
This figure illustrates the progression of accuracy across epochs for five iterations, highlighting improvements and comparisons between each iteration. 
Figure b: Loss Trends Across Epochs for Multiple Iterations
This figure depicts the reduction of loss across epochs for five iterations, demonstrating the convergence behavior and performance stability over time.}
    \label{fig:fig11}
\end{figure}

\bigskip
\section{CONCLUSIONS}\label{sec:Conclusions}
This study demonstrates a comprehensive approach to addressing the challenge of missing information in Depth images, a critical aspect of Autonomous Driving Systems (ADS). By employing iterative training combined with deep learning techniques, we successfully generated and refined Depth images from single RGB inputs, achieving a notable accuracy improvement from 82.78\% to 90.19\% over five iterations. The iterative refinement process not only enhanced the accuracy but also significantly reduced the percentage of missing pixels in Depth images.

Additionally, leveraging the U-Net algorithm for image segmentation allowed us to automate and accelerate the correction process, further improving the prediction accuracy to 96.44\%. These advancements were validated using the Cityscapes dataset, which served as an effective benchmark for urban scene understanding in autonomous driving applications. Our methodology demonstrated robust performance in filling missing information, as evidenced by substantial improvements in corrected pixel percentages and accuracy metrics.

This work lays a strong foundation for future research aimed at enhancing Depth image generation and refinement techniques. The iterative training approach and segmentation-based corrections can be extended to other datasets and use cases, such as 3D reconstruction, robotics, and other computer vision applications, where accurate depth information is paramount. Future directions may include optimizing the computational efficiency of the model and exploring multi-modal input strategies to further improve depth prediction performance in real-time scenarios.

\bigskip
\bibliographystyle{apalike}
{\small
\bibliography{bibliography}}

\appendix
\section*{\appendixname \ A}\label{Appendix A}
We calculate the average number of black pixels across all Depth images in the Cityscapes dataset using the following formula:

    \[Average\ black\ pixels = \frac{1}{N}\sum_{P\in \mathcal{P}}^{N} \Pi \]
Where $\mathcal{P}$ is the pixels in the image, $ N$ is the number of images, which is 150 thousand images in our work and $\Pi$ as the following:
\begin{equation*}
\Pi = 
\begin{cases} 
    1 & \text{if }  p = 0 \\ 
    0 & \text{else } 
\end{cases}
\end{equation*}

\section*{\appendixname \ B}\label{Appendix B}
In our work, we employed an NVIDIA RTX 4090 graphics card, an Intel Core i9-14900K CPU, and 64 GB of RAM operating at 4000 MT/s.

\end{document}